# Feedback System Neural Networks for Inferring Causality in Directed Cyclic Graphs

By: William Schoenberg (University of Bergen, Norway & isee systems inc. Lebanon NH, USA)

**Abstract**
This paper presents a new causal network learning algorithm (FSNN, Feedback System Neural Network) based on the construction and analysis of a non-linear system of Ordinary Differential Equations (ODEs). The constructed system provides insight into the mechanisms responsible for generating the past and potential future behavior of dynamic systems. It is also interpretable in terms of real system variables, providing a wholistic, causally accurate, and systemic understanding of the real-life interactions governing observed phenomena.  This paper demonstrates the generation of an n-dimensional ordinary differential equation model that can be parameterized to fit measured data using standard numerical optimization techniques.  The model makes use of feed forward artificial neural nets to capture nonlinearity, but is a parsimonious and interpretable representation of the network of causal relationships in complex systems.  The generated model can easily and rapidly be experimented with and analyzed to determine the origins of behavior using the loops that matter method (Schoenberg et. al 2019).  A demonstration of the utility and applicability of the method is given, showing that it produces an accurate, and causally correct model for a three state, non-linear, complex dynamic system of known origin. Generalization to other dynamic systems with other data sources is then discussed.

**Introduction:**
Successfully identifying the true causal relationships in complex systems that produce a certain dynamic behavior has arguably been a main focus of science throughout time.  As science evolves and technology improves, the methods and tools we use to discover causality ought to change and adapt to keep pace.  Present-day machine learning methods, including probabilistic modeling, kernel machines or deep learning, have arisen from a strong, almost single minded, focus on data-oriented empiricism making use of the extraordinary amounts of observational data available from a plethora of sources (Ghahramani, 2015; Schölkopf & Smola, 2008; Goodfellow et. al., 2016).  When viewed through the lens of accurate prediction power, machine learning has proven to be quite successful but, present-day machine learning techniques typically fail to reveal the fundamental causal mechanisms driving behavior.  Although, it must be said that interpreting the structure behind black box, deep learning models is an active research area (Montavon et. al, 2018).  To make full use of the technological advancements in machine learning, significant emphasis must be placed on finding a valid and interpretable causal understanding of the underlying real-world system (Runge et. al, 2019).

Observational causal inference as a field started with applied statistics and has grown out of the seminal works of Wiener and Granger (Wiener, 1956; Granger, 1969).  The most well-known method for observational causal inference is Granger causality which tests whether omitting the past of a time series $X$ in a time series model including $Y$'s own and other covariates' past increases the prediction error of the next time step of $Y$ (Granger, 1969).  Granger causality is useful in discovering specific causal links in a system, but it fails to generalize to complex non-linear systems, typically failing to identify all of the links in the networks of feedback relationships which govern these systems from a wholistic perspective (Spirtes & Zhang, 2016).

Non Granger methods in observational causal inference can be categorized into the following three broad categories as done in the literature by Runge et. al (2019): nonlinear state-space methods, causal network learning algorithms and structural causal model frameworks.  Of particular interest are the techniques which underlie causal network learning algorithms, especially the PC algorithm and its brethren.  These methods start with either an empty network or a fully connected network and iteratively add or remove edges testing against an invariant payoff function to evaluate the likelihood of the validity of the causality of the generated network structures (Verma & Pearl, 1990; Spirtes & Glymour, 1991; Zhang, 2008; Runge et. al, 2018; Runge, 2018).  These methods are of particular interest because they start by generating a structure which then generates a behavior.  Methods of this class have the potential for the broadest of applications and are particularly relevant to systems which are representable as ordinary differential equations (ODEs), because networks of causal relationships are simply, and as Forrester argues, ideally represented (and therefore analyzed) in the form ODEs (1994).

The major problem with the state of the art in the field of observational causal inference is that it does not work in feedback rich systems which by definition cannot be represented as directed acyclic graphs.  This method demonstrates a new approach to causal network learning algorithms which generates then analyzes an n-dimensional, non-linear, complex system of ODEs to do automated causal inference with a high level of accuracy from a wholistic systems perspective incorporating feedback and time delays.  This approach provides insight and understanding into the mechanisms responsible for generating past and future behavior of dynamic systems, giving users a wholistic, causally accurate, and systemic understanding of the real-life networks governing observed phenomena.

**The method**
At the center of a FSNN (Feedback System Neural Network) is an attempt to apply the core principles of empirically driven machine learning to observational causal inference.  Key to the function of this method is the construction of a model which is a highly non-linear complex system of ODEs evaluated over a time dimension.  The system of ODEs, referred to as a generated model, is analyzed to determine the origins of behavior.  In the generated model, the state variables constitute the memory of that model and their relationships (through perceptron firing) are the source from which behavior originates.  To build the generated model, first the method constructs the system of ODEs which links the state variables together via their derivatives.  Second via a process of optimization the generated model is parameterized to reproduce the observed state trajectories the system exhibited.  Third, the now parameterized generated model is studied and the mathematical relationships between the state variables are objectively measured which clarifies the origins of behavior within the empirically trained, generated model.  Finally, the causal explanation is then extracted and can be validated by subject matter experts, turning the fruits of the method from a black box data-driven machine learning model into a transparent and analyzable structure-driven model.

The first step in creating the generated model is to generate a system of ODEs to be parameterized. To do that the state variables in the ground truth system are enumerated and initialized from observed data. Next, in the model being generated, the derivative of each state variable is set to be a function of every other state variable as well as itself. It is these relationships in the derivative functions that represent the opportunity for the learning that will take place during the model training/parameterization process. The process of constructing the derivatives for each state variable offers the end-user the ability to add a-priori knowledge to the model by specifying which causal relationships are known not to exist, and therefore can be excluded from the learning process. This is done by eliminating the relationship between any directed pair of source and target state variables.

The relationships between each directed pair of state variables can take any form that the user desires. There are generally two intervention areas available. The first is the form of the equation of the term(s) that will represent each state variable in the derivative of each other state variable including itself. The second, is the form of the equation to combine the terms representing each state variable in the derivative function of all state variables. The factors to consider in making the decision of what formulations to use are the general applicability of the range of producible behavior modes to the observational data being matched and the computational complexity in terms of the numbers of parameters necessary to train the generated model while being mindful of overspecification and degenerate payoff surfaces.

The optimal structure discovered to parameterize and train the model is one which connects all the state variables via a neural network as is done in the hidden state of neural ODEs (Chen et. al, 2018). For the purposes of this method, the states are not a part of the hidden network. Specifically, for each state variable in the model, there is a single multilayer perceptron neural net constructed such that there is one output, which is the derivative for the state variable, and the input layer is constructed using the current value of all the state variables at each calculation interval. The hidden state of each of these neural networks is individually constructed according to best practice and the specific dataset being studied. This structure is ideal because neural nets are universal approximators, theoretically allowing them to learn any non linear function which dictates the relationships between state variables (Cybenko, 1989). The universal approximation theorm allows this meta-model to be applicable to any ground truth system. To specify that a certain relationship between state variables does not exist, simply remove the input node representing the source from the derivative function of the target's neural net's input layer.

The second step is model training/parameterization and is where an optimization process is run defining as a payoff function the minimization of the squared error between the training data and the calculated values for each state variable. Any ODE solver can be used to simulate the model, and in the example a Runge-Kutta 4 (RK4) solver was used. The optimization algorithm used in the example case was Powell's BOBYQA (bound optimization by quadratic approximation) as implemented in the public open source project DLib version 19.7.

The third step, once the trained model has been generated, is to analyze that generated model using the link score metric from the loops that matter method to measure and extract the meaningful causal relationships (Schoenberg et. al, 2019). The link score metric reports the contribution at a point in time that an independent variable (x) has on a dependent variable (z) arranged in the function $z = f(x, y)$. For this approach the link score is used to measure the impact that one state variable has on another so that the causal relationship between state variables can be quantified and understood. Equation (1) is reproduced from Schoenberg et. al. below. Link scores are computed using the calculation interval employed by the ODE solver.

The link score for the link x → z is:

$$LS(x \to z) = \begin{cases} \left( \left| \frac{\Delta_x z}{\Delta z} \right| \cdot sign\left( \frac{\Delta_x z}{\Delta x} \right) \right), & \\ 0, & \Delta z = 0 \text{ or } \Delta x = 0 \end{cases}$$

(1)

Where $\Delta z$ is the change in z from the previous time to the current time. $\Delta x$ is the change in $x$. $\Delta_x z$ is the change in $z$ with respect to $x$. From a computational perspective $\Delta_x z$ is the amount $z$ would have changed, conditionally, if $x$ had changed the amount it did, but $y$ had not changed. The first major term in this equation represents the magnitude of the link score, the second is the link score polarity.

The magnitude of the effect (force is a good analogy) that x has on z is relative to all of the effects on z. This is a dimensionless quantity, and if all of the effects are in the same direction, it is the fraction of the change in z that originates in a change in $x$. If the formulation of $z$ is linear, then the values are restricted to the range [0,1]. When there are negative and positive effects, these numbers may be very large in magnitude, but this does not harm the overall analysis of the generated model (Schoenberg, et. al 2019). The absolute value is used because the change in z could be in either direction due to the effects from other variables, regardless of the magnitude of the effect that $x$ has, implying that the polarity can and would be wrong.

The polarity of a link is defined as the sign of the partial difference at time t. This formulation is the same as the one used in Richardson (1995), though the formulation there was as a partial derivative, not difference. The polarity numerator is the same as it is for the magnitude, but the denominator is the change in $x$. When $x$ does not change, the score is by definition 0, though the magnitude would be 0 in any case since no change in $z$ would be attributable to $x$.

Link scores can be multiplied together following the chain rule of partial differentiation allowing the calculation of the system of equations to proceed at any desired aggregation level (that is with more or fewer intermediate algebraic computations). Typically for models generated using this technique link scores are calculated along the pathway of the relationship from one state variable directly to the derivative of another state variable. It is possible to calculate link scores which pass through other state variables back to the original state variable forming complete feedback loops. This is valuable for models with known or physically bound structure, but of less value for all but the simplest of generated models. This is because this method of generating differential equations produces (assuming no a-priori information), a number of feedback loops which is the factorial of the number of state variables, numbers which quickly overwhelm computational and analytic evaluations.

When following the standard practice of calculating link scores which measure the impact of one state variable (source) on the derivative of another (target), the generated link scores are then normalized across all of the relationships which have the same target, at each point in time. This produces a signed percentage score which describes specifically and objectively what percentage of the behavior of the target state variable is contributed by the source state variable, including what the polarity of their relationship is at that specific point in time.

**Application to a non-linear dynamic system**

To properly demonstrate the application of a FSNN, a model of a three-state system was used as the ground truth. The ground truth system is an ODE which consists of three state variables ($State_1, State_2, State_3$) where each state variable is directly connected to itself, and its predecessor in a large feedback loop and each state is connected to itself with a short feedback loop. This system contains 4 feedback loops in total (all with a balancing polarity), and the equations are shown in Table 1. The training datasets were generated by simulating the ground truth system using an RK4 solver with a $dt$ of 1/4 for 100 time units producing three full cycles of a dampened oscillation shown in Figure 1. The training data was sampled at the end of each time unit, giving 100 time points per state variable across two distinct initializations of the states. Initialization 1 (29, 96, 4), Initialization 2 (22, 11, 78).

Table 1: Equations for ground truth system with units. $f$ is a monotonically increasing sigmoid function over the x range [0,100] and y range [0, 100] with inflection point at (50, 50).

| | |
|---|---|
| $State_1 = flow_2 - flow_1$ | Units |
| $State_2 = flow_3 - flow_2$ | Units |
| $State_3 = flow_4 - flow_3$ | Units |
| $flow_1 = (g - f(State_3))/t$ | Units/Time |
| $flow_2 = State_1/t$ | Units/Time |
| $flow_3 = State_2/t$ | Units/Time |
| $flow_4 = State_3/t$ | Units/Time |
| $t = 5$ | Time |
| $g = 75$ | Units |
| $f$ | Non-linear sigmoid function |

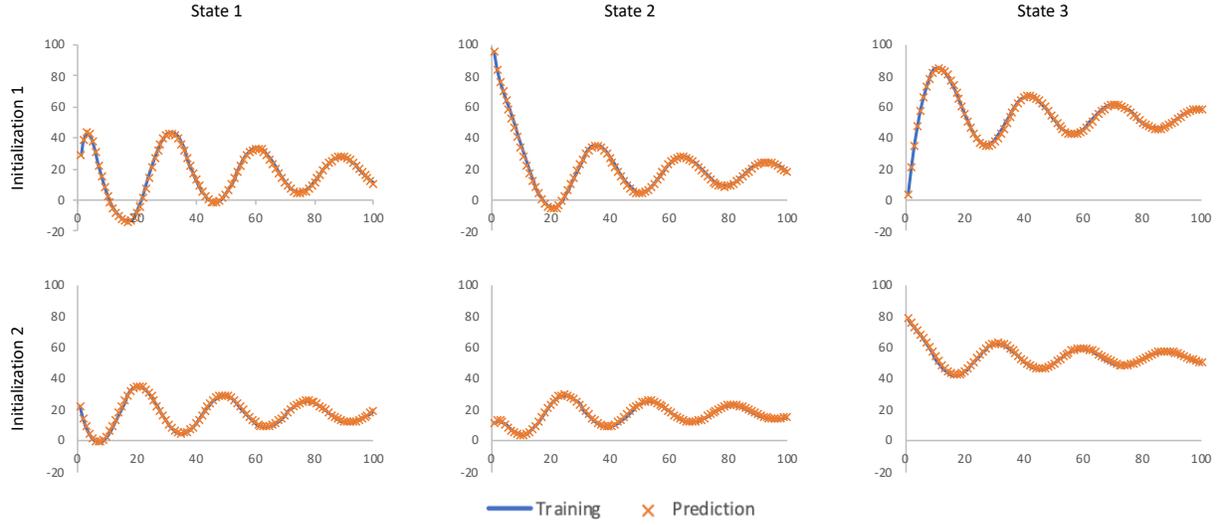

*Figure 1: Performance of the generated model on the two training initializations.  Initialization #1 (29, 96, 4), Initialization #2 (22, 11, 78).*

The training progressed in a single phase where all network weights and bias parameters were set to 0 representing a starting point with no causal structure.  Each neural net was constructed with three hidden layers, of 8 then 6 then 4 nodes, and each neural network took input from all three states and produced a single output.  The activation function used in all nodes was a $tanh$ function.  All state variables used as inputs were linearly re-scaled to the range [-1, 1] by dividing the value of each state variable by 100 (an externally determined maximum potential magnitude for any of the states) and multiplying the value of the output node by 100 so that the state variable values would remain accurate to the ground truth system.  Figure 1 demonstrates that the model performed well on the two training datasets, where the blue lines are the data generated by the ground truth system and the orange points are the predictions by the generated model.

To validate that the generated model is a causally accurate recreation of the ground truth system, the generated model must reproduce the proper causal connections as measured in the source model, which is demonstrated in Figure 2.  A pathway based link score analysis of the first training set initialization of the ground truth system was performed, measuring the polarity, and nature of the relationship between each state variable and each other state variable including itself.  The results of that analysis are the blue lines in Figure 2.  Negative numbers mean that the relationship has a negative polarity, and are therefore balancing, positive numbers mean a positive polarity, and are therefore reinforcing.  The magnitude of the numbers represents the strength of the connection at each point in time.  The same analysis was then performed on the generated model using the same first training set initialization.

An analysis of Figure 2 yields that conclusion that the generated model has identified the proper causal links in the ground truth system, and that the generated model has properly identified the polarity for each of those links.  Figure 2 also indicates that the generated model is not the singular well defined model of the ground truth system because the exact nature of the relationships doesn't follow the exact same pattern.

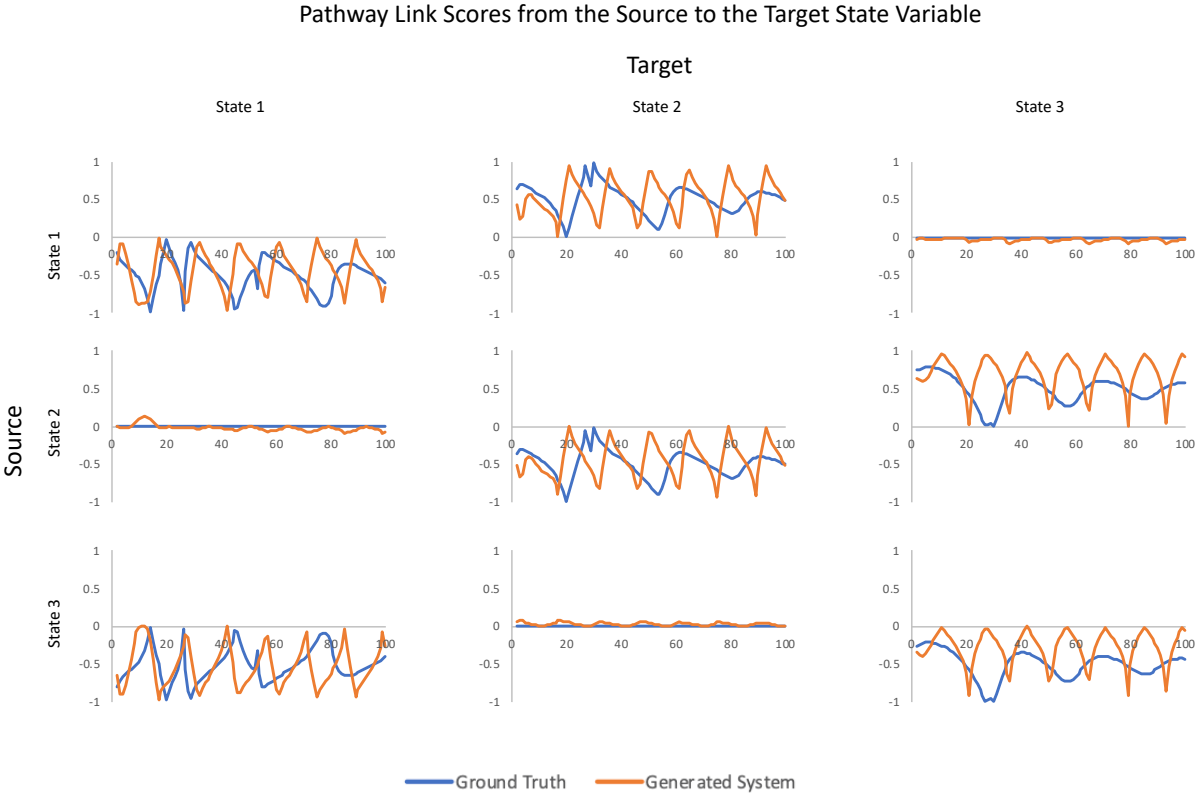

*Figure 2: Demonstration of identification of proper causal links and their associated polarities. Graphs showing 0 magnitudes means there is no connection between the source and target. Negative values mean a balancing link, positive values a reinforcing link. The magnitude represents the strength of the connection at that instant in time.*

Next, the generated model's performance on untrained, testing data was measured. Figure 3 demonstrates the results of the error distribution across 100 unique Monte Carlo generated, random initializations using Sobol sequences (Sobol et. al, 1999). Sobol sequences were used to best explore the selected input state space (Burhenne et. al, 2011). The selected input state space was limited to the condition that the sum of the initial value of the states must be within the range [30 – 150]. The distribution of that total initial value to each of the three states was unrestricted. This condition is based upon the sum of the initial states in the two initializations used to train the model, this important decision is explained in depth below in the next section. Figure 3 demonstrates remarkable performance in reproducing the behavior of the ground truth system in the large majority of the test cases. The 95% confidence bounds of the error for all three states are close to 0 which demonstrates that the generated model is an accurate recreation of the ground truth system in the behavioral space under the given range of initial conditions.

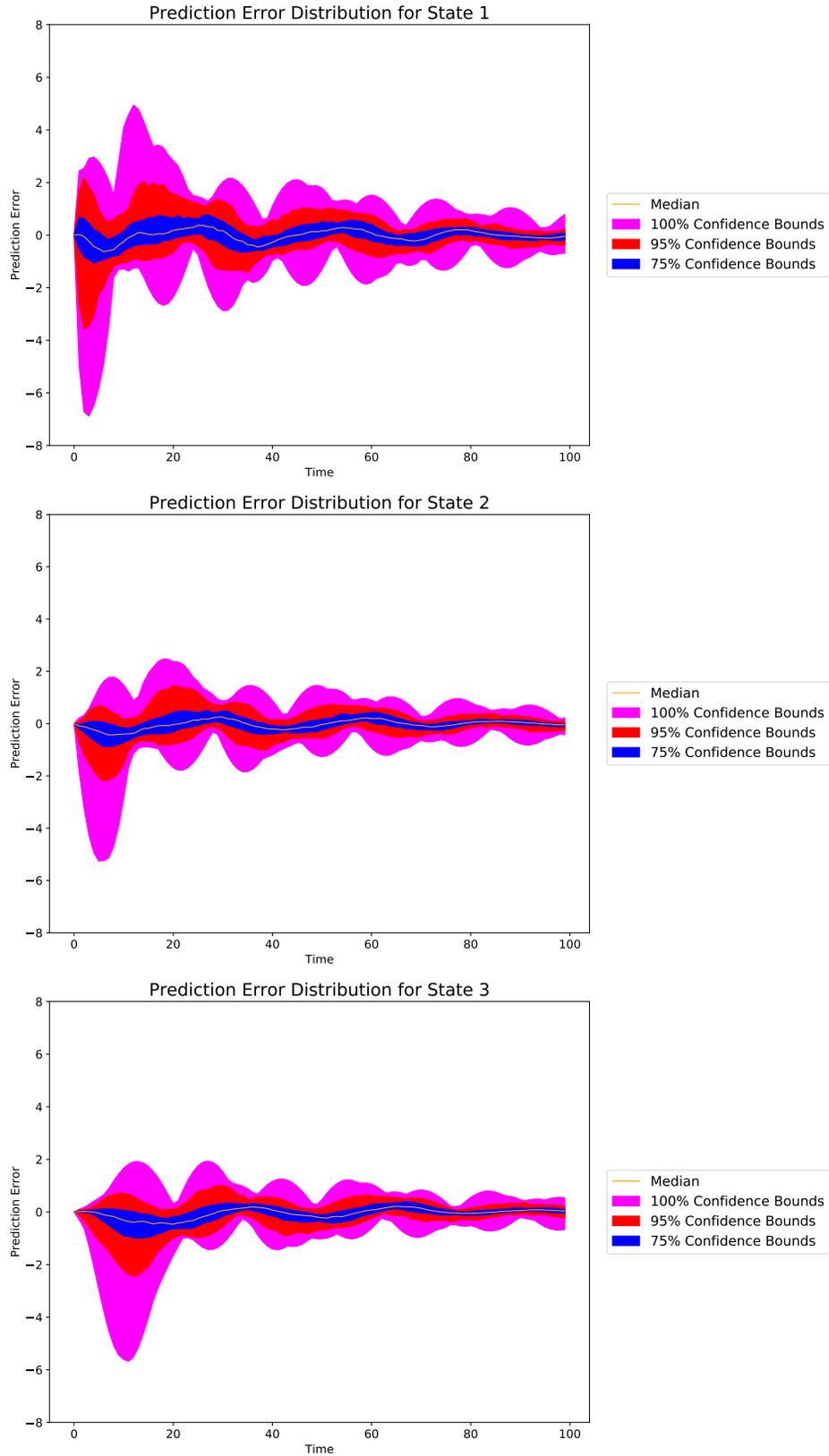

*Figure 3: Prediction error for each of the three states from 100 Monte Carlo generated initializations of the generated model where the sum of the initial state values was between 30 and 150.*

**Challenges and considerations for dealing with non linear feedback systems**

The above example demonstrates the problems with finding the singular well-defined model which perfectly explains the ground truth system in the structural space, and therefore as a result in the behavioral space as well. The generated model above was not the singular well-defined model for the ground truth system, because the link score profiles seen in Figure 2 did not match perfectly in all time points for all relationships. Instead, the generated model was one of a set of models which matches the behavior over a range of initializations well, because in the structural space it shares many similarities with the ground truth system. Because the true structure of the ground truth system was known, it was therefore determinable that the recreation of that ground truth system did not match in every way, but it was possible to confirm that the recreation matched the key, high level attributes of the causal structure including: identification of all correct causal links and their polarities, plus the general forms for the relationships governing the ground truth system's behavior. The consequence is, not surprisingly, that the model successfully passes the behavior validation tests to which it was exposed.

Even after infinite training on infinite data, the meta-model at the heart of this method may not be well defined for a specific ground truth system, and the chances of finding that singular well defined model falls quickly as the amount of training, or data is cut to finite levels. Because a neural network is a universal approximator (Cybenko, 1989), the meta-model is capable of producing an infinitely wide range of behavior patterns, encompassing an infinitely large set of possible models. After training, the generated model is just one possible explanation, a simple hypothesis for the causal relationships governing the origins of the given data. That explanation may just as easily be proven wrong by the introduction the next piece of data which encoded within it are more hints about the subtle facets of the structure of the ground truth system which created it. Finding the singular well-defined model of the ground truth system in a synthetic data experiment like the one presented here is a fool's errand, which, even if possible, potentially requires infinite training on infinite data because of the unbounded range of the ground truth system. For many systems, the distinction between the generated behaviors of the singularly well-defined model of the system, and a (potentially quite large) set of highly accurate generated models of that same system, is miniscule to insignificant in the behavioral space, but vastly different in the structural (i.e. causal) space. Specifically, the causal structures behind those two classes of models can potentially be totally unrelated, sharing very little, if anything at all in common. At the end of the day, the only way to establish with any certainty real ground truth in unknown systems is via validation of and empirical experimentation on, the hypothesized causal structures, far beyond what any behavior-based validation regime can do.

The example analyzed in this paper is a demonstration of the method under a favorable set of conditions. In this experiment, the generated model could be trained across two distinct unrelated initializations, with full, perfectly accurate (no measurement error) and regular data available for all system states with no extraneous data. In addition, the magnitude for the system states was well-defined, and the training data sampled well the potential behavioral space of the testing data, while producing a well sloped payoff surface. The results of this experiment would have been different if any one part of this perfect storm of conditions was not present.

The best way to understand the construction and training of the generated model is to imagine the ground truth system, and therefore the generated model as an $n$-sized set of $n$-dimensional functions, where $n$ is the number of state variables of the ground truth system being studied. Each one of those $n$-dimensional functions describes, irrespective of time, the functional relationship between all the states and the change in each specific state. Each one of those $n$-dimensional functions produces a manifold in state space, whose surface is the unique fingerprint that yields the information necessary to identify each of the $n$-dimensional functions which are the causal structure of the system. During training the goal is to examine as much of each manifold as possible so that the neural nets whose job it is to learn each of those $n$-dimensional functions can reproduce as much as is possible of the ground truth manifolds in the generated model. Each additional point in time that is used in training brings along with it a new opportunity to examine another portion of each manifold.

Training across multiple distinct initializations helps to combat overfit, expands the range of the generated model, and reduces problems with confounders by exposing the generated model to more facets of the manifolds of the ground-truth system. Multiple initializations better exercise the ground truth system, revealing additional information which would not be accessible from just a single simulation. Initializing the system in multiple distinct ways tends to break any false learned causations in the generated model, more commonly referred to as confounders. This is because, while in one initialization two state variables may exhibit similar state trajectories because of the impact by a third one, in a second initialization, the first and second states may exhibit opposite state trajectories hinting at a proper identification of state three as a driver of one and two. Even with just two distinct initializations, performance significantly improves regarding the identification of the proper causal structure because the incidence of confounders drops precipitously. This is because the chances of repeated correlations across any two specific timesteps in the first derivatives of the state variables drops.

Increasing the number of initializations trained on, increases the scope of the generated model because, during the training phase, the generated model explores more of the manifolds of the ground truth system. When the number of initializations used in training is low, only a few of the potential combinations of the values taken by the $n$ state variables are actually encountered. Consequently, only a very small fraction of the total surface area of the $n$ manifolds is actually explored and exploited in the identification of the underlying causal structure. Therefore, each neural network derived is generally only applicable to a small fraction of the full (potential) input space because the function which describes the relationships between the states at those unexplored values is still completely unknown.

Each unique initialization constitutes a combination of values from the $n$ state variables and may produce a new $n$ dimensional state trajectory. Each single initialization tends to settle into a specific area of the manifolds generated by the ground truth system (with the exception of chaotic systems), therefore adding more samples to any one initialization tends to be less important.  To demonstrate, all forms of non-linear dynamic systems at the limit of time can be categorized by their behavior.  There are three such categories of systems.  First there are chaotic systems which never settle into a regular behavior pattern.  Second there are oscillatory systems, which are systems that at their limit produce a standing oscillation through the same sets of values.  Finally, there are systems in equilibrium, which over time tend towards producing state variables which do not change.  In all but the chaotic systems, eventually with enough time spent simulating, for each initialization a point will be reached where no additional portions of the manifolds are being explored because all combinations of state variables which are producible by that initialization have already been encountered.  Therefore, in all but chaotic systems each new initialization brings with it more opportunity to find new never before seen combinations of the state variables, therefore providing insight on more of the manifolds of the ground truth system, and therefore improving the performance of the generated model.  This is why each additional initialization of the ground truth system is so important for training purposes versus each additional sample of an already explored initialization.

For the case of the specific example above, the surface of the ground truth system's manifolds was not explored where multiple states took on values of 70 or greater.  This is because the ground truth system always produces a dampened oscillation, and training took place with an initial sum of the states of 129, and 111.  Therefore, as is demonstrated in Figure 4, when the initial sum of the states is greater than 150 the prediction begins to rise exponentially.  It does this because at initialization values of 150 or greater, the probability of encountering multiple states which simultaneously have values over 70 rises.  The prediction error gets exponentially worse as the sum of the initial states rises because the probability of encountering further unexplored areas of the manifolds significantly and quickly increases.

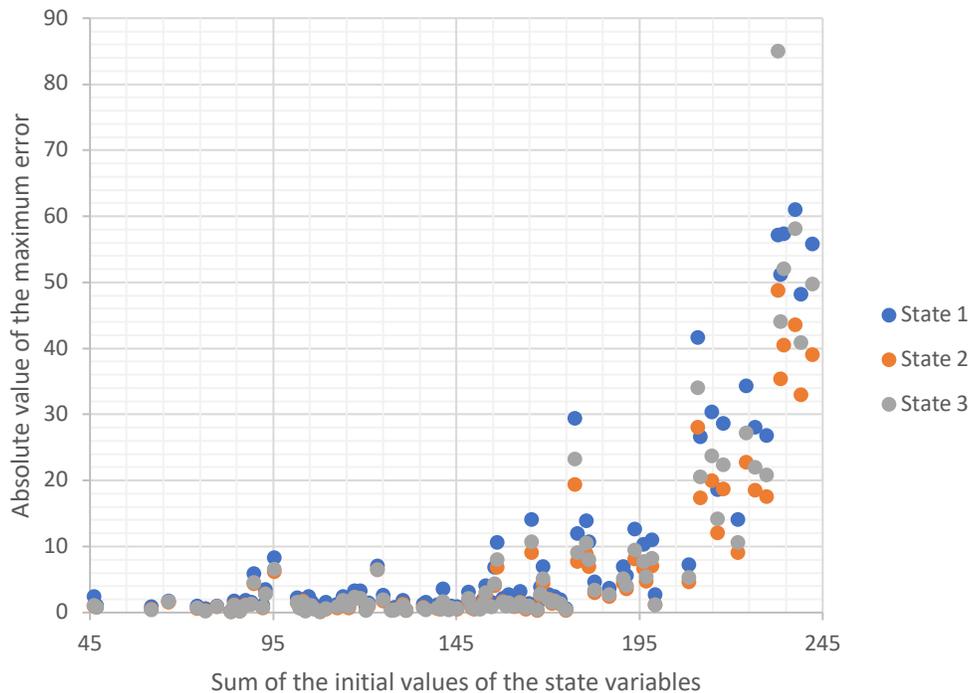

*Figure 4: The absolute value of the maximum observed error distribution from 100 Monte Carlo generated, random initializations, using Sobol sequences (Sobol et. al, 1999). Sobol sequences were used to best explore the selected input state space (Burhenne et. al, 2011). This figure shows how leaving the training area adversely affects the generated model's utility, because the training has not been sufficient to discover the non linear functions governing the inter-relationships between the states in areas where there are two or more states which have relatively large values (+70) simultaneously. The generated model is effective over the range of 0 – 150 in the sum of the initial values of the 3 state variables.*

When training across multiple initializations, it is important to verify that the initializations are indeed distinct, and that each set of initial states does not show up as a set of calculated values in any of the other training initializations, otherwise the initializations are linked (one initialization is the direct outcome of another), and there is no potential to see additional facets of the manifolds of the ground truth system structure because new state trajectories are not being produced. In real-world systems with an unknown ground truth, this process can be recreated by studying multiple unrelated instances of the system where each instance must share the same exact causal structure.

The utility of the method in situations with anything less than perfect data is still unexplored. Real world systems are messy, and it yet needs to be studied how the method handles measurement error, irregularity in collected data, missing data, and the incorporation of extraneous data. There are many significant challenges in this space still. Because the method always fits to, and therefore attempts to explain all given time-series data endogenously, any error in specification of either extraneous time series data, or measurement error in the data, will cause problems with learning the true causal relationships  The design of the method forces an endogenous explanation for all time-series data it trained on, including measurement error, and unrelated inputs.  Both forms of error will be endogenized into the system, and therefore be made part of the feedback structure of the system, leading to improper identification of causal structure.  Specifically, extraneous, causally unexplainable time-series inputs will cause the shape of the payoff surface to warp towards degeneracy, because the extraneous data will not be explainable by any of the given state variables and yet the method by its very design and construction must search for an endogenous explanation for its origins.

The final piece of key information known in the example case was the magnitude of the state variables.  Because of the use of the $tanh$ activation function, a linear rescaling of all state variables was done.  This specific combination of a linear rescale, and $tanh$ function causes problems with signal attenuation, although, these problems will be present in any combination of a normalization process with the application of a non-linear activation function.  In this specific case, as the absolute value of any state variable grows above its associated magnitude, the behavior is pushed into the far reaches (domain) of the $tanh$ function, where it planes out, reducing the marginal impact significantly and quickly.  For example, functionally, the system responds in nearly the same way to a state variable which has a value two times the associated magnitude vs the same state variable having a value of that magnitude.  The ramifications of this problem are especially important for time series forecasting, especially in systems with exponential data distributions because of the right skew in observable values which is fundamentally incompatible with a linear scaling.  This problem is even more insidious because setting larger and larger magnitudes then reduces the marginal impact of changes typically observed on the left side of the data distribution.  Therefore, it appears that a logarithmic scaling may be called for in systems with exponentially distributed time series data.  Regardless, further study is warranted to understand the relationship between activation function, the scaling function, and the time-series data distribution.

**Conclusions**
Even in the face of what may seem like insurmountable problems, this line of work, and within that scope, the FSNN in particular are of great value.  This specific approach represents a significant improvement in our ability to undertake causally accurate science, using less human input, and has the potential to underlie a revolution in the way that causal structures are identified.  For a moment, imagine that significant progress has been made addressing all the challenges mentioned above, and this method is functionally able to identify very nearly the singularly well-defined model with correct structural underpinnings among the large majority of ground truth systems under relatively achievable data requirements.  In that case, this method becomes the basis for an artificial intelligence, surpassing that of human beings, which predicts and explains with a high level of accuracy never observed phenomena which occur deep into the future.  As often said, "...with great challenges come great rewards...".